\documentclass[sigconf]{acmart}
\usepackage{amssymb}
\usepackage{booktabs}
\usepackage[T1]{fontenc}
\usepackage{textcomp}
\usepackage[scaled=.92]{helvet}
\usepackage{graphicx}
\usepackage{balance}
\usepackage{ccicons}
\usepackage{ragged2e}
\usepackage{subcaption}
\usepackage{amsmath}
\usepackage{algorithm}
\usepackage[noend]{algpseudocode}

\begin{document}
\copyrightyear{2017}
\acmYear{2017}
\setcopyright{rightsretained}
\acmConference{MobiCom '17}{October 16--20, 2017}{Snowbird, UT, USA}
\acmDOI{10.1145/3117811.3131269}
\acmISBN{978-1-4503-4916-1/17/10}

%\CopyrightYear{2017}
%\setcopyright{rightsretained}
%\conferenceinfo{MobiCom '17}{October 16--20, 2017, Snowbird, UT, USA}
%\isbn{978-1-4503-4916-1/17/10}
%\doi{https://doi.org/10.1145/3117811.3131269}

%Authors, replace the red X's with your assigned DOI string. See pdf attached to ACM rightsreview confirmation email.

\fancyhead{}
\settopmatter{printacmref=false, printfolios=false}

\title{Poster: DeepTFP: Mobile Time Series Data Analytics based Traffic Flow Prediction}
%\subtitle{Extended Abstract}

%=================================================================================================================================================================

\author{Yuanfang Chen, Falin Chen, Yizhi Ren, Ting Wu, Ye Yao}
\affiliation{
    \institution{Cyberspace School, Hangzhou Dianzi University}
    \city{Hangzhou}
    \country{China}
    }
\email{yuanfang.chen.tina@gmail.com, cyuanfang@acm.org}
\email{0linsrz@gmail.com}
\email{{renyz, wuting, yaoye}@hdu.edu.cn}

\renewcommand{\shortauthors}{Yuanfang Chen et al.}

%=================================================================================================================================================================

\begin{abstract}
Traffic flow prediction is an important research issue to avoid traffic congestion in transportation systems.  Traffic congestion avoiding can be achieved by knowing traffic flow and then conducting transportation planning.  Achieving traffic flow prediction is challenging as the prediction is affected by many complex factors such as inter-region traffic, vehicles' relations, and sudden events.  However, as the mobile data of vehicles has been widely collected by sensor-embedded devices in transportation systems, it is possible to predict the traffic flow by analysing mobile data.  This study proposes a deep learning based prediction algorithm, DeepTFP, to collectively predict the traffic flow on each and every traffic road of a city.  This algorithm uses three deep residual neural networks to model temporal closeness, period, and trend properties of traffic flow.  Each residual neural network consists of a branch of residual convolutional units.  DeepTFP aggregates the outputs of the three residual neural networks to optimize the parameters of a time series prediction model.  Contrast experiments on mobile time series data from the transportation system of England demonstrate that the proposed DeepTFP outperforms the Long Short-Term Memory (LSTM) architecture based method in prediction accuracy.
\end{abstract}

%==============================================================================================================================================================




\maketitle

%=================================================================================================================================================================

\section{Introduction}
Traffic flow prediction is an important research issue, and it is a feasible method to be able to solve or alleviate some serious problems in an Intelligent Transportation System (ITS), for example, energy saving and emission reduction problems caused by traffic congestion.  Traffic congestion is a serious problem in the traffic management of big cities across the world.  Moreover, the public safety caused by traffic congestion is another serious problem.  For example, in a study of 471 U.S. urban areas of 2014~\cite{scorecard2015}, the extra energy cost due to traffic congestion was \$160 billion (3.1 billion gallons of fuel).  In addition, long periods of traffic congestion force the release of more carbon dioxide (CO$_{2}$) greenhouse gases into the atmosphere, and increase the number of accidents.  Predicting citywide traffic flow patterns can help reduce traffic congestion by planning and controlling the traffic, and therefore reduce the amount of CO$_{2}$ emissions as well as save lives.

It is possible to predict the citywide traffic flow by analysing mobile time series data, because of the traffic flow correlation between different traffic roads, as shown in Figure~\ref{fig:traffic_flow}.
\begin{figure}[!ht]
  \centering
  \includegraphics[width=2.5in]{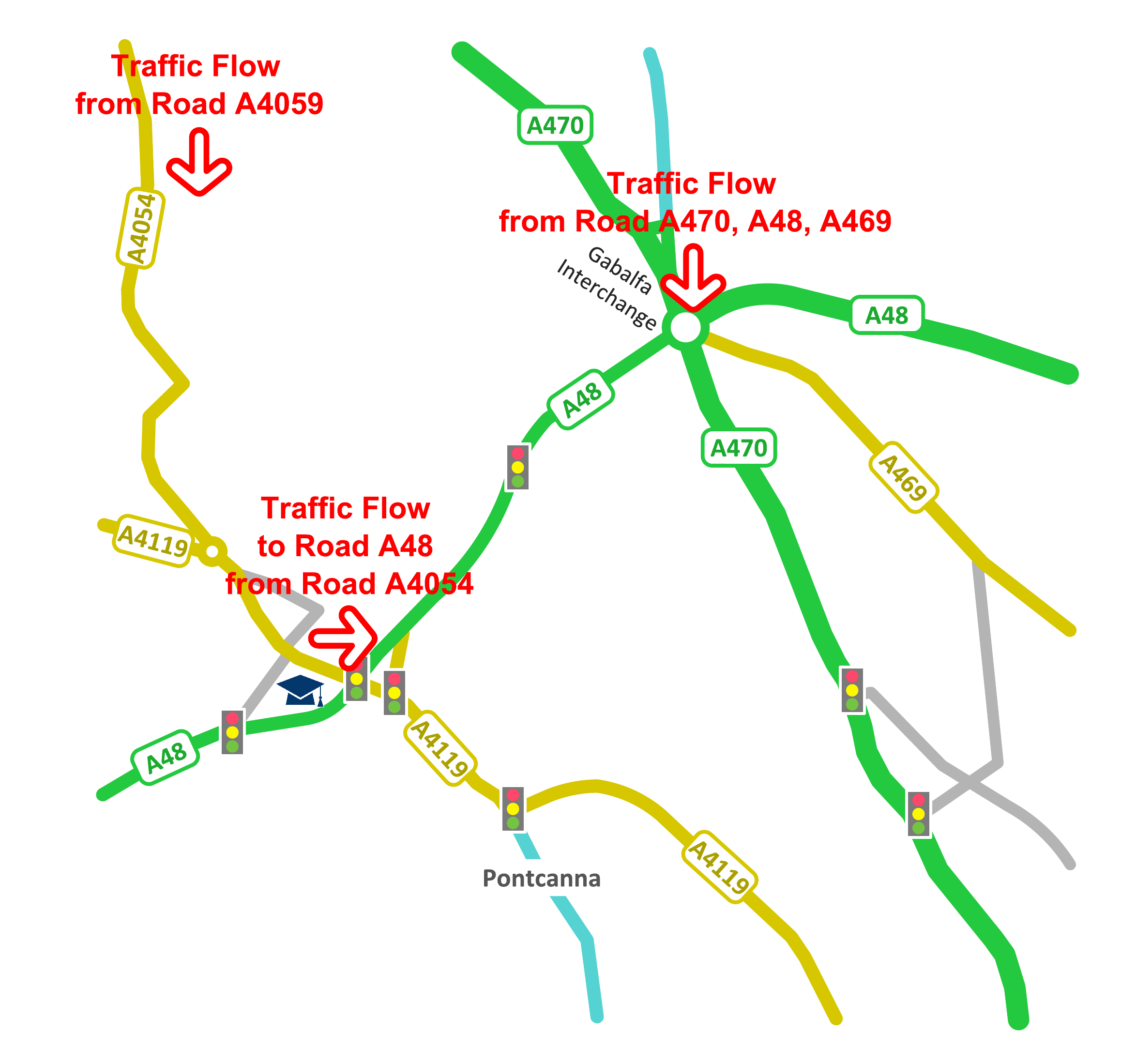}\\
  \caption{Traffic flow relationship between different traffic roads.}
  \label{fig:traffic_flow}
\end{figure}

The traffic flow of a traffic road is the total number of vehicles driving on the road during a given time interval.  It is estimated by counting the passing vehicles with various wireless devices, e.g., Automatic Number Plate Recognition (ANPR) cameras, in-vehicle Global Positioning System (GPS), and inductive loops built into road surfaces.  The spatio-temporal relevance of the traffic flow between different traffic roads makes the prediction possible.

Many kinds of techniques have been proposed to address the traffic flow prediction problem in different situations, and traditionally there is no single best method for every situation.  It is better to combine several suitable techniques to improve the accuracy of prediction under considering different situations. It means that the traditional traffic flow prediction methods are not able to satisfy most real-world application requirements.  During the last five years, some studies have tried to use data analytics to solve the traffic flow prediction problem, and the results demonstrate that such schemes are feasible and are able to improve the accuracy of prediction, as the method proposed in~\cite{lv2015traffic}.  Such kind of data analytics based prediction method is able to satisfy the requirements of different applications by analyzing the data from each corresponding specific application.

This paper proposes an algorithm DeepTFP to predict the traffic flow of a city.  DeepTFP consists of two modules: (i) deep learning module.  It designs three residual neural networks to model the temporal closeness, period, and trend properties of citywide traffic flow; (ii) time series function module.  DeepTFP dynamically aggregates the outputs of the three residual neural networks to optimize the parameters of the time series prediction function $\hat{X}_{t+1}=c+\varepsilon_{t+1}+\sum\limits_{i=1}^{n}\theta_{i}X_{t}$.

%=================================================================================================================================================================

\section{DeepTFP}
DeepTFP consists of two parts: (i) residual neural network framework, and (ii) parameter optimized time series model.

Figure~\ref{fig:traffic_flow} illustrates the architecture of DeepTFP.
\begin{figure}[!ht]
  \centering
  \includegraphics[width=2.9in]{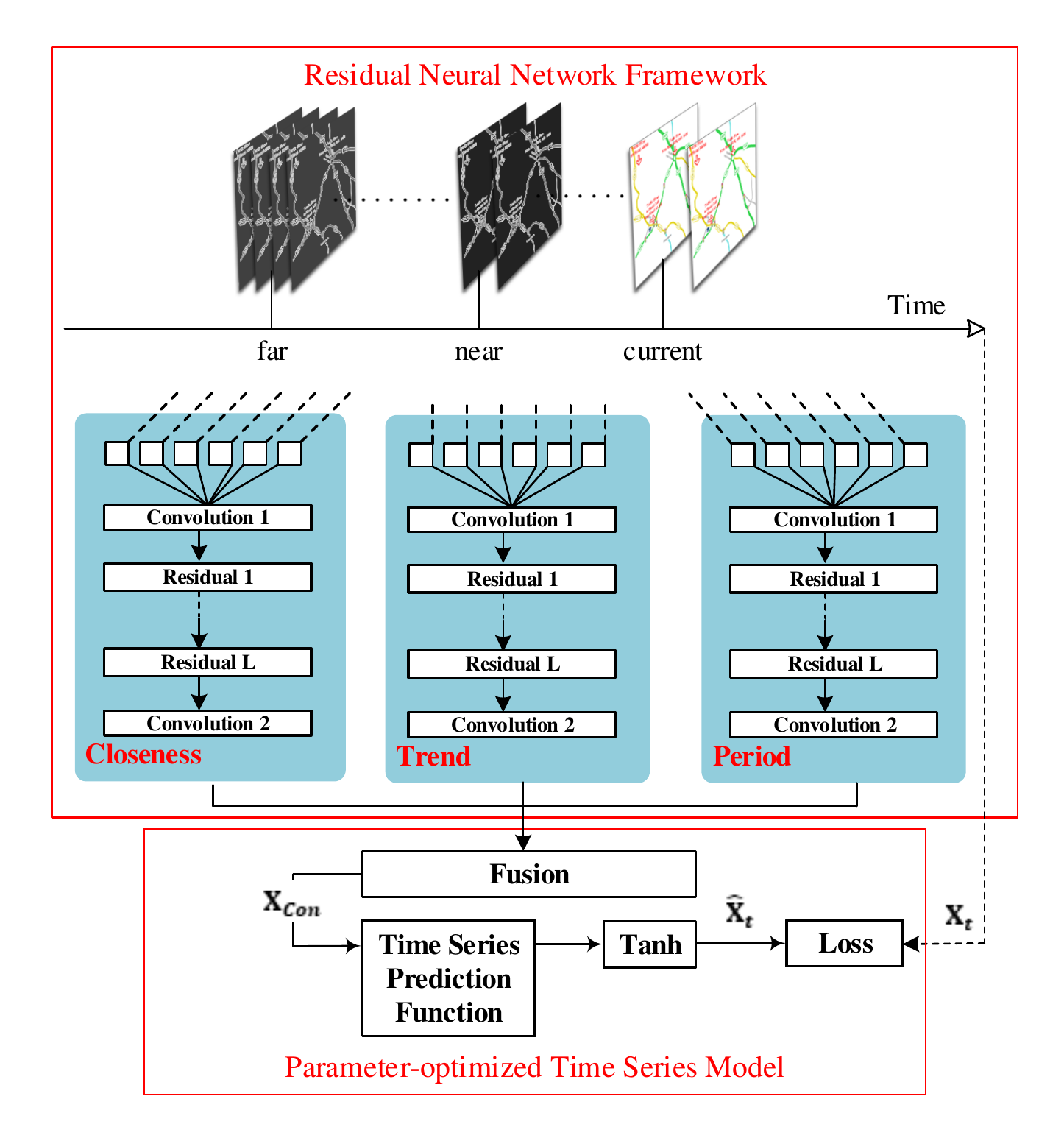}\\
  \caption{Architecture of DeepTFP.}
  \label{fig:traffic_flow}
\end{figure}

DeepTFP has three components to model the temporal closeness, trend, and period properties of the traffic flow from different roads.  The three components have the same network structure.  Such structure captures the spatial dependency between traffic roads.  To capture the spatial dependency of connecting traffic roads, a residual neural network with convolutions and residual units is designed.  The convolution is formulated as: $Con_{c/t/p}^{(1)}=f(W_{c/t/p}^{(1)}\ast Con_{c/t/p}^{(0)}+b_{c/t/p}^{(1)})$, where $*$ denotes the convolutional process, $f$ is an activation function, e.g., the rectifier $f(z):=max(0,z)$, and $W_{c/t/p}^{(1)}$, $b_{c/t/p}^{(1)}$ are the parameters of the convolution.

The outputs of the three components are fused as $X_{Con}$, and the outputs are used as the input of the time series prediction function to be used to optimize the parameters of the function by minimizing the mean squared error between the predicted flow values and the true flow values: $\mathcal{L}(\theta)=||X_{t}-\hat{X}_{t}||_{2}^{2}$, where $\theta$ denotes the set of learnable parameters of the time series prediction function $\hat{X}_{t+1}=c+\varepsilon_{t+1}+\sum\limits_{i=1}^{n}\theta_{i}X_{t}$.

Algorithm~\ref{alg:deeptfp} presents the steps of DeepTFP.
\begin{algorithm}[!ht]
\caption{Deep Learning based Traffic Flow Prediction}
\label{alg:deeptfp}
\begin{algorithmic}[1]
\State \textbf{Input:} Time series observations: $\{X_{1}, ..., X_{n}\}$; lengths of closeness, period, trend sequences: $l_{c}, l_{p}, l_{q}$; period $p$; trend span: $q$.
\State \textbf{Output:} Predicted traffic flow value $X_{t+1}$.
\Statex $\vartriangleright$ construct training instances
\State $\mathcal{D}\leftarrow\emptyset$
\For{\texttt{all available time interval t($1 \leq t \leq n$)}}
\State \texttt{$\mathcal{S}_{c}=[X_{t-l_{c}}, ..., X_{t}]$}
\State \texttt{$\mathcal{S}_{p}=[X_{t-l_{p}\cdot p}, ..., X_{t-p}]$}
\State \texttt{$\mathcal{S}_{q}=[X_{t-l_{q}\cdot q}, ..., X_{t-q}]$}
\State put an training instance ($\{S_{c}, S_{p}, S_{q}\}, X_{t}$ into $\mathcal{D}$
\EndFor
\Statex $\vartriangleright$ train the model
\State initialize all learnable parameters $\theta$ in the time series prediction function
\Repeat
\State randomly select a batch of instances $\mathcal{D}_{b}$ from $\mathcal{D}$
\State input $\mathcal{D}_{b}$ into residual convolutional units to get $\hat{X}^{'}_{t}$
\State input $\hat{X}^{'}_{t}$ into the time series prediction function as the $X_{t}$ to get $\hat{X}_{t}$
\State find $\theta$ by minimizing the objective $\mathcal{L}(\theta)=||X_{t}-\hat{X}_{t}||_{2}^{2}$ with $\mathcal{D}_{b}$
\Until{stopping criteria is met}
\Statex $\vartriangleright$ predict the traffic flow value $X_{t+1}$
\State input the traffic flow $X_{t}$ into trained time series prediction function
\State return $\hat{X}_{t+1}$
\end{algorithmic}
\end{algorithm}

%=================================================================================================================================================================

\section{Evaluation Results}
This section compares the performance of DeepTFP and Long Short-Term Memory (LSTM) architecture based method.

LSTM is a type of recurrent neural network~\cite{gers2000learning}.  It is capable of learning order dependence in sequence prediction problems.

\subsection{Experimental Data}
\begin{figure}[!ht]
  \centering
  \includegraphics[width=3.3in]{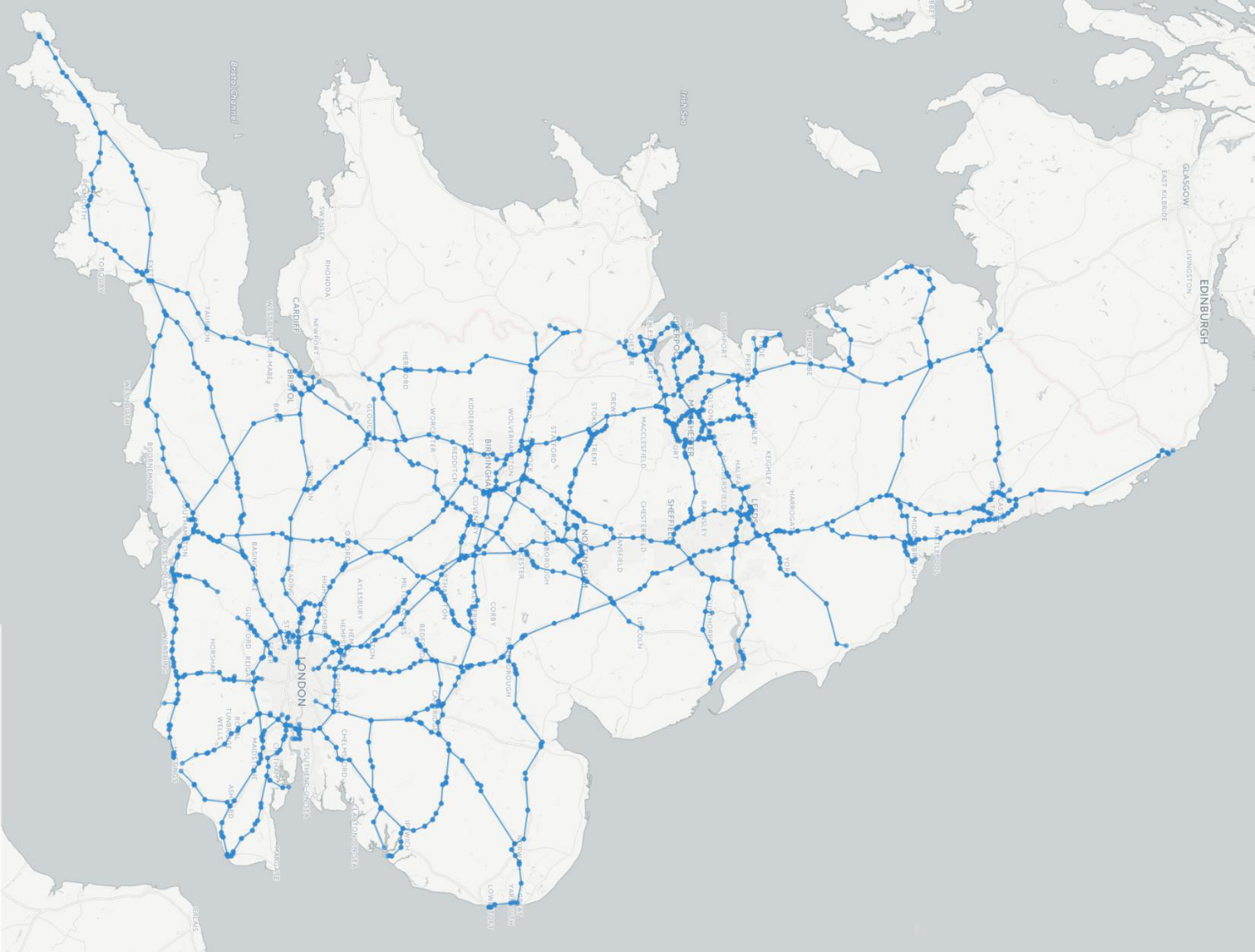}\\
  \caption{Flow information of traffic roads is acquired by analyzing the trajectory data of the vehicles driving on these traffic roads.  The update period of this information is 15 minutes.  Highways and arterial roads are marked by blue lines and dots, where the dots denote the beginnings and ends of roads.}
  \label{fig:data_description}
\end{figure}

In this study, trajectory data of vehicles is used, which is a kind of series transportation data.  This data can be used to calculate and predict the traffic flow of a transportation system.  It is collected from the transportation system of England.  This transportation system has 2501 traffic roads\footnote{A two-way highway/arterial road is counted as two traffic roads.} covering 300 miles of England highways and arterial roads, which is illustrated in Figure~\ref{fig:data_description}.

\subsection{Comparative Results}
Comparative experiments measure and compare average prediction results.  Each line is the average of 2501 traffic roads and 31 days of one month.

The Root-Mean-Square Error (RMSE)~\cite{chai2014root} is used to measure the differences between the values actually observed and the values predicted by methods.  It is defined as the square root of the mean square error: RMSE$(\hat{\theta})=\sqrt{MSE(\hat{\theta})}=\sqrt{E((\hat{\theta}-\theta)^{2})}$, where $\theta$ is the values actually observed, and $\hat{\theta}$ is used to denote the values predicted by a method.

\begin{figure}[!ht]
    \centering
    \begin{subfigure}[b]{0.5\textwidth}
        \includegraphics[width=\textwidth]{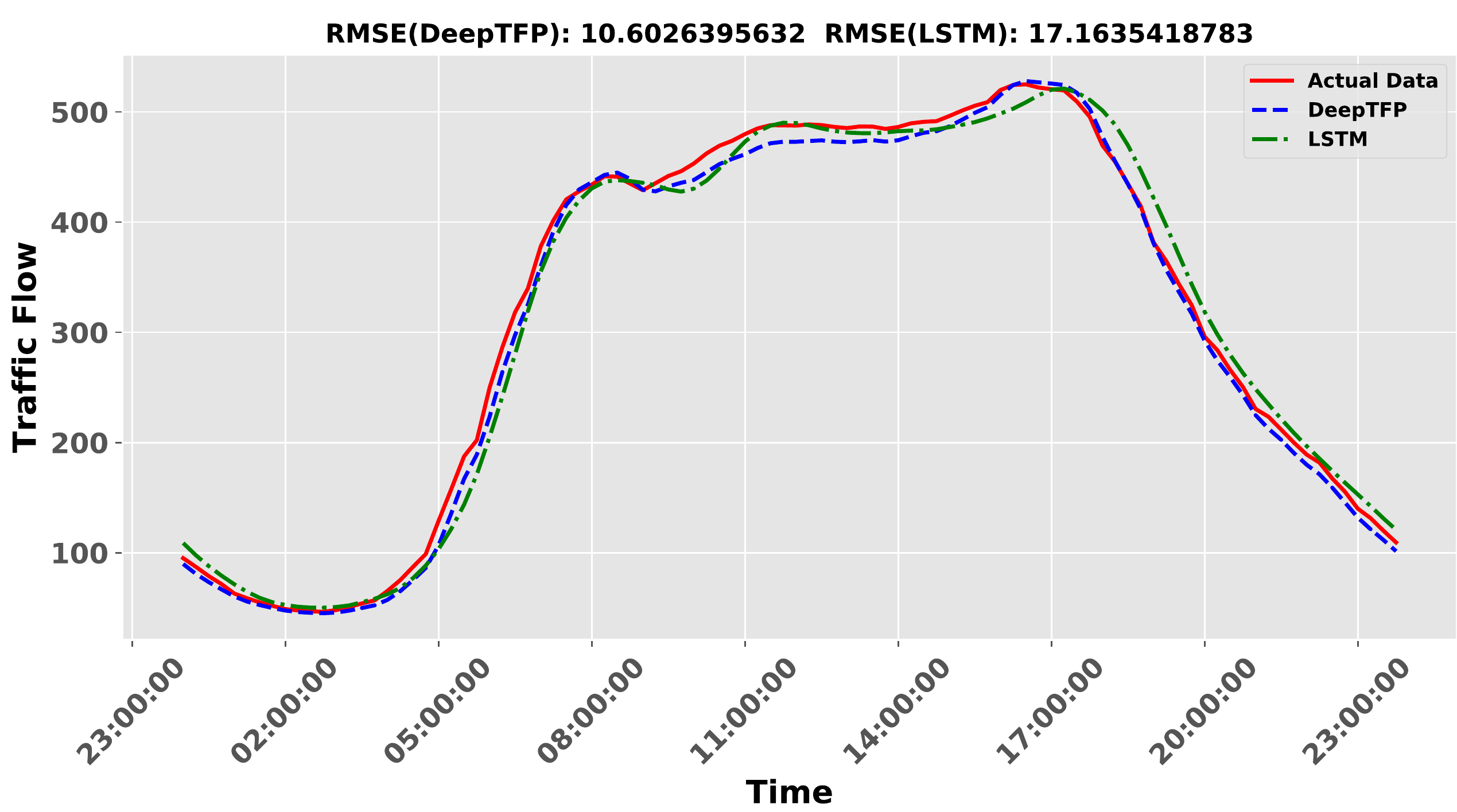}
        \caption{Two months of training data (October and November) are used to train the model.  One month (December) is predicted with the trained model.}
        \label{fig:comparative_results_average_two_month}
    \end{subfigure}
    \begin{subfigure}[b]{0.5\textwidth}
        \includegraphics[width=\textwidth]{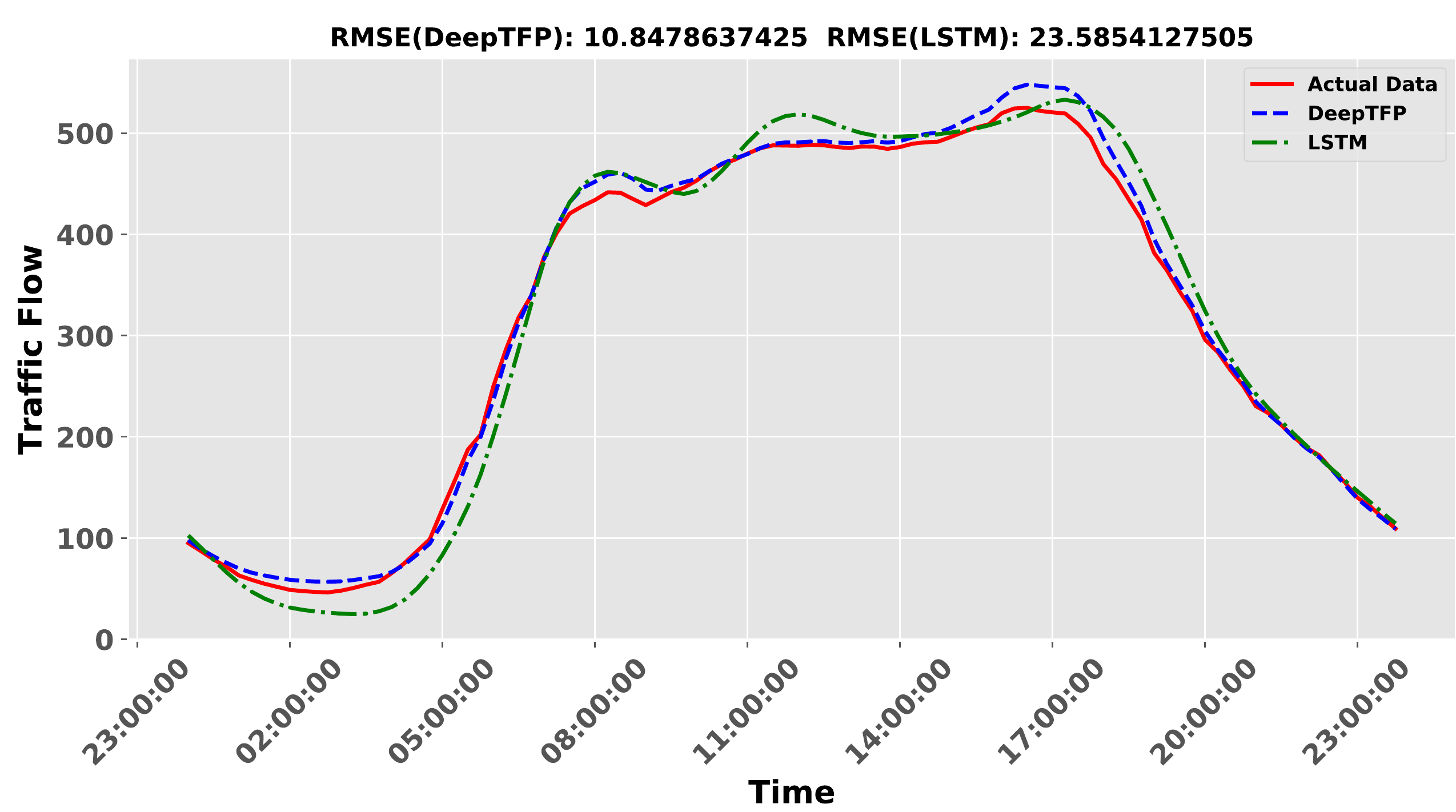}
        \caption{One month of training data (November) is used to train the model.  One month (December) is predicted with the trained model.}
        \label{fig:comparative_results_average_one_month}
    \end{subfigure}
    \caption{Predicted values with DeepTFP and the LSTM architecture based prediction method, and comparative results with actual values.}
    \label{fig:comparative_results_average}
\end{figure}

The results are illustrated in Figure~\ref{fig:comparative_results_average}.

These are two important observations which can be used to direct algorithm design to improve the performance of a prediction algorithm:
\begin{enumerate}
  \item The performance of prediction does be improved by training the model with the actual data;
  \item Increasing the volume of training data is able to improve the performance of prediction.  The deep learning framework is effective for organizing high-volume data to train a prediction model.
\end{enumerate}

%=================================================================================================================================================================

\bibliographystyle{ACM-Reference-Format}
\bibliography{sample}

%=================================================================================================================================================================

\end{document}